\documentclass[5pt]{article}

\usepackage[letterpaper]{geometry}
\usepackage{spconf,amsmath,epsfig}
\usepackage{enumitem}
\usepackage{hyperref} 

\usepackage{amsmath}
\usepackage{amssymb}
\usepackage{graphicx}
\usepackage{subfigure}
\usepackage{booktabs}
\usepackage{multirow}
\usepackage{array}
\usepackage{color}

\newcolumntype{C}[1]{>{\centering}p{#1}}

\usepackage{fancyhdr}
\begin{document}\sloppy

% Example definitions.
% --------------------
\def\x{{\mathbf x}}
\def\L{{\cal L}}

% Title.
% ------
\title{Knowledge-Guided Recurrent Neural Network Learning \\ for Task-Oriented Action Prediction}
%
% Single address.
% ---------------
\name{Liang Lin, Lili Huang, Tianshui Chen, Yukang Gan, and Hui Cheng\thanks{\noindent L. Huang and T. Chen share equal authorship. This work was supported by State Key Development Program under Grant 2016YFB1001004 and National Natural Science Foundation of China under Grant 61622214.}}
\address{School of Data and Computer Science, Sun Yat-Sen University, Guangzhou, China \\
\small{linliang@ieee.org, \{huanglli3, chtiansh, ganyk\}@mail2.sysu.edu.cn, chengh9@mail.sysu.edu.cn, }}
%, tianshuichen@gmail.com, ganyk@mail2.sysu.edu.cn, chengh9@mail.sysu.edu.cn, lianglin@ieee.org
% For example:
% ------------
%\address{School\\
%	Department\\
%	Address\\
%   Email}
%
% Two addresses (uncomment and modify for two-address case).
% ----------------------------------------------------------
%\twoauthors
%  {A. Author-one, B. Author-two\sthanks{Thanks to XYZ agency for funding.}}
%	{School A-B\\
%	Department A-B\\
%	Address A-B}
%  {C. Author-three, D. Author-four\sthanks{The fourth author performed the work
%	while at ...}}
%	{School C-D\\
%	Department C-D\\
%	Address C-D\\
%   Email}
%

\maketitle

\begin{abstract}
This paper aims at task-oriented action prediction, i.e., predicting a sequence of actions towards accomplishing a specific task under a certain scene, which is a new problem in computer vision research. The main challenges lie in how to model task-specific knowledge and integrate it in the learning procedure. In this work, we propose to train a recurrent long-short term memory (LSTM) network for handling this problem, i.e., taking a scene image (including pre-located objects) and the specified task as input and recurrently predicting action sequences. However, training such a network usually requires large amounts of annotated samples for covering the semantic space (e.g., diverse action decomposition and ordering). To alleviate this issue, we introduce a temporal And-Or graph (AOG) for task description, which hierarchically represents a task into atomic actions. With this AOG representation, we can produce many valid samples (i.e., action sequences according with common sense) by training another auxiliary LSTM network with a small set of annotated samples. And these generated samples (i.e., task-oriented action sequences) effectively facilitate training the model for task-oriented action prediction. In the experiments, we create a new dataset containing diverse daily tasks and extensively evaluate the effectiveness of our approach.
\end{abstract}

\section{Introduction}
\begin{figure}[!t]
   \centering
   \includegraphics[width=1.0\linewidth]{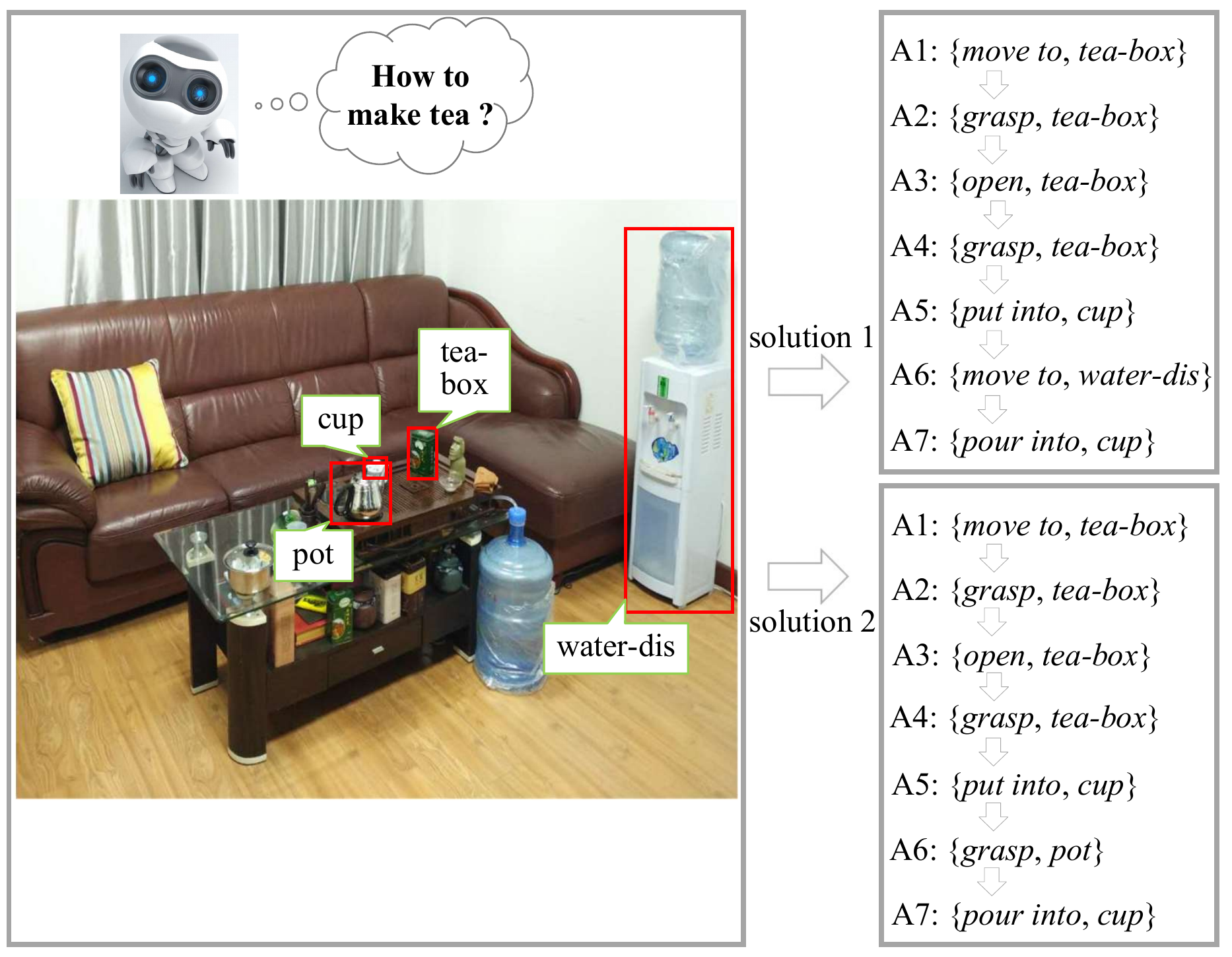}
   \caption{Two alternative action sequences for completing the task ``make tea" under a specific office scene, which are inferred according to joint understanding of the scene image and task semantics. An agent can achieved this task by successively executing either of the  sequences.}
   \label{fig:task-definition}
   \vspace{-12pt}
\end{figure}

Automatically predicting and executing a sequence of actions given a specific task would be one quite expected ability for intelligent robots~\cite{thrun2005probabilistic}. For example, to complete the task of  ``make tea" under the scene shown in Figure \ref{fig:task-definition}, an agent need to plan and successively execute a number of steps, e.g., ``move to the tea box", ``grasp the tea box". In this paper we aim to train a neural network model to enable such a capability, which was rarely addressed in computer vision and multimedia research.

We regard this above discussed problem as task-oriented action prediction, i.e., predicting a sequences of atomic actions towards accomplishing a specific task. And we refer an atomic action as a primitive action operating on an object, denoted in the form of two-tuples $A=(action, object)$. Therefore, the prediction of action sequences depends on not only the task semantics (i.e., how the task to be represented and planned) but also the visual scene image parsing (e.g., recognizing object categories and their spatial relations in the scene). Since recent advanced deep convolutional networks (CNNs) achieve great successes in object categorization and localization, in this work we assume that objects are correctly located in the given scene. However, this problem remains challenging due to the diversity of action decomposition and ordering, long-term dependencies among atomic actions, and large variation of object layout in the scene.

We develop a recurrent long-short term memory (LSTM) \cite{hochreiter1997long} network to harness the problem of task-oriented action prediction, since LSTM models has been demonstrated their effectiveness on capturing long range sequential dependencies, especially for the tasks like machine translation \cite{cho2014learning} and image captioning \cite{vinyals2015show}. These approaches usually adopt the encoder-decoder architecture, in which an encoder  first encodes the input data (e.g., an image) into a semantic-aware feature representation and a decoder  then decodes this representation into the target sequence (e.g., a sentence description). In this work, we interpret the input image into a vector that contains the information of object categories and locations and feed it into the LSTM network (named Action-LSTM) with the specified task, and the network is capable of generating the action sequence through the encoder-decoder learning.

In general, it usually requires large amounts of annotated samples to train LSTM networks, especially for tackling the complex problems such as task-oriented action prediction. To overcome this issue, we present a two-stage training method by employing a temporal And-Or graph (AOG) representation~\cite{lin2009stochastic,xiong2016robot}. First, we define the AOG for task description, which hierarchically decomposes a task into atomic actions according to their temporal dependencies. In this semantic representation, an and-node represents the chronological decomposition of a task (or sub-task); an or-node represents the alternative ways to complete the certain task (or sub-task); leaf-nodes represent the pre-defined atomic actions. The AOG can thus contain all possible action sequences for each task by embodying the expressiveness of grammars. Specifically, given a scene image, a specific action sequence can be generated by selecting the sub-branches at all of the or-nodes with a Depth-First Search (DFS) manner. Second, we train an auxiliary LSTM network (named AOG-LSTM) to predict the selection at the or-nodes in the AOG, and can thus produce a large number of new valid samples (i.e., task-oriented action sequences) that can be used for training the Action-LSTM. Notably, training the AOG-LSTM requires only a few manually annotated samples (i.e., scene images and the corresponding action sequences), because making selection in the context of task-specific knowledge (represented by the AOG) is seldom ambiguous. 

The main contributions of this paper are two-folds. First, we raise a new problem called task-oriented action prediction and create a benchmark (including 13 daily tasks and 861 RGB-D images captured from 16 scenarios)\footnote{For more details please refer to \url{http://hcp.sysu.edu.cn/}.}. Second, we propose a general approach for incorporating complex semantics into the recurrent neural network learning, which can be generalized to various high-level intelligent applications.

\vspace{-8pt}
\section{Related work}
We review the related works according to two main research steams: task planning and recurrent sequence prediction. 

\noindent{\bf Task planning.} In literature, task planning (aslo referred to symbolic planning~\cite{sung2013learning}) has been traditionally formalized as the deduction \cite{allen1991planning} or satisfiability \cite{kautz1992planning} problems for a long period. Sacerdoti et al. \cite{sacerdoti1974planning} introduced hierarchical planning, which first planned abstractly and then generated fine-level details. Yang et al. \cite{yang2007learning} utilized the PDDL representation for actions and developed an action-related modeling system to learn an action model from a set of observed successful plans. Some works also combined symbolic with motion planning \cite{kambhampati1991combining}. Cambon et al. \cite{cambon2009hybrid} regarded symbolic planning as a constraint and proposed a heuristic function for motion planning.
Plaku et al. \cite{plaku2010sampling} extended the work and planned with geometric and differential constraints. Wolfe et al. \cite{wolfe2010combined} proposed a hierarchical task and motion planning algorithm based on the hierarchical transition networks. Although working quite well in the controlled environments, these methods required encoding every precondition for each operation or domain knowledge, and they could hardly generalize to the unconstrained environments with large variance \cite{sung2013learning}. Most recently, Sung et al. \cite{sung2013learning} represented the environment with a set of attributes, and proposed to use the Markov Random Field to learn the sequences of controllers to complete the given tasks.

\noindent{\bf Recurrent sequence prediction.} Recently, the recurrent neural networks has been widely used in various sequence prediction tasks, including natural language generation \cite{hochreiter1997long}, machine translation \cite{cho2014learning}, and image captioning \cite{vinyals2015show}. These works adopted the similar encoder-decoder architecture for solving sequence prediction. Cho et al. \cite{cho2014learning} mapped the free-form source language sentence into the target language by utilizing the encoder-decoder recurrent network. Vinyals et al. \cite{vinyals2015show} applied the similar pipeline for image captioning, which utilized a CNN as the encoder to extract image features and an LSTM network as the decoder to generate the descriptive sentence.

\begin{figure*}[!t]
\centering
\subfigure[]{
\includegraphics[width=0.48\linewidth]{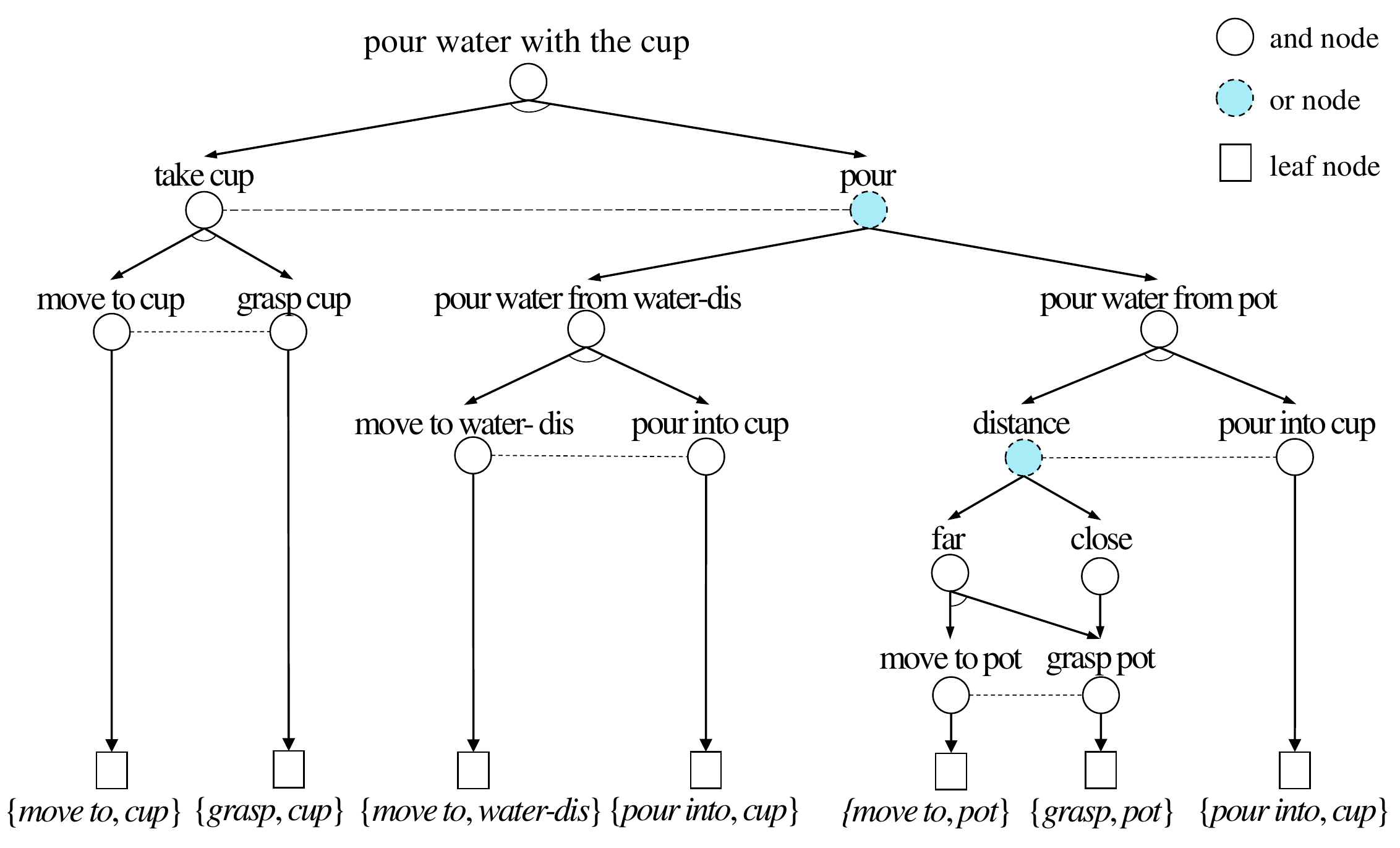}}
\subfigure[]{
\includegraphics[width=0.50\linewidth]{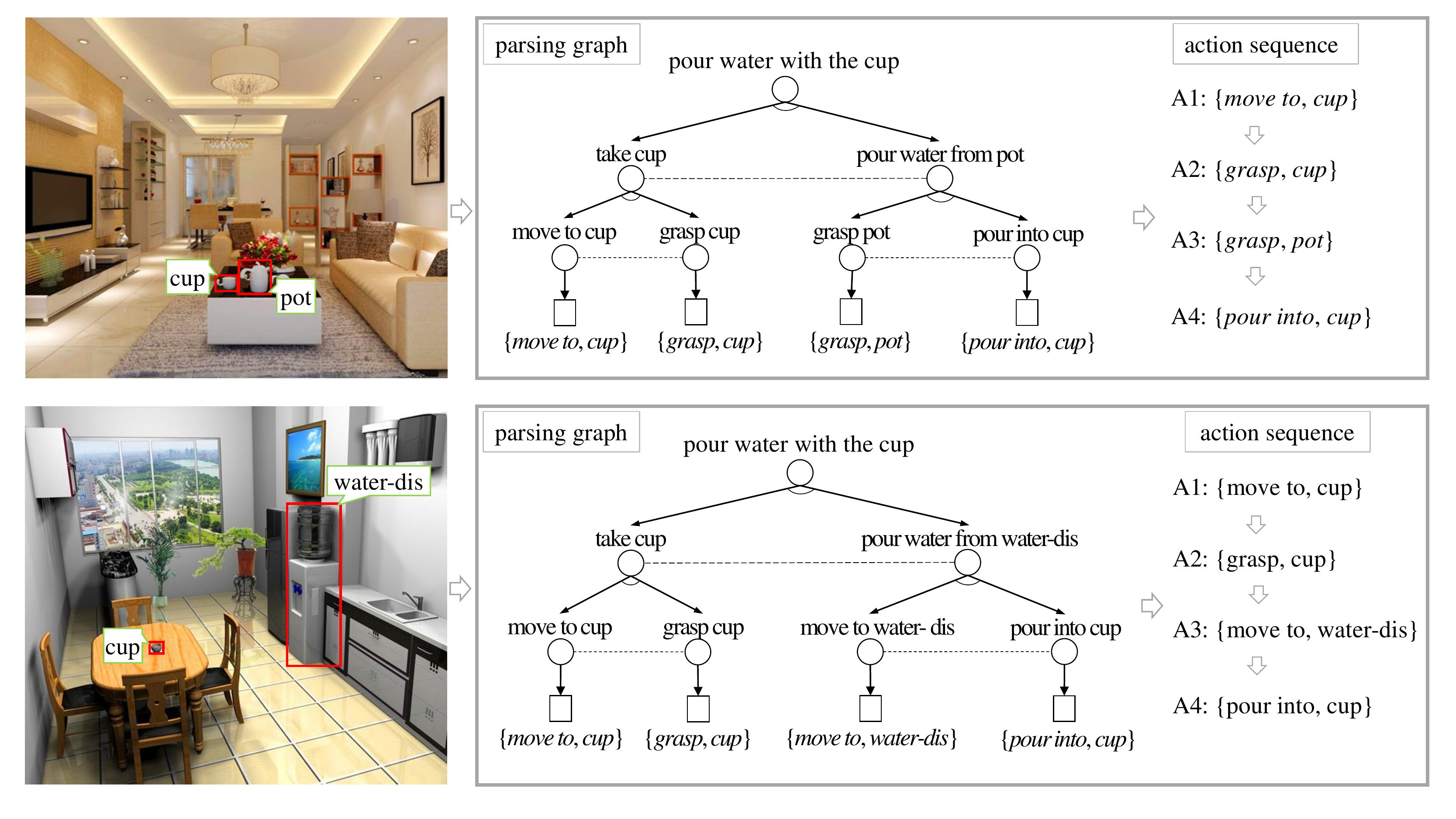}}
\vspace{-8pt}
\caption{An example of temporal And-Or graph for describing the task ``pour water with the cup"  shown in (a) and two parsing graphs and their corresponding action sequences under two specific scenes shown in (b).}
\label{fig:aog}
\vspace{-6pt}
\end{figure*}

\vspace{-8pt}

\section{LSTM background}
We start by briefly introducing the technical background of recurrent LSTM networks. The LSTM is developed for modeling long-term sequential dependencies. In addition to the hidden state $\mathbf{h}_t$, the network contains an extra memory cell $\mathbf{c}_t$, input gate $\mathbf{i}_t$, forget gate $\mathbf{f}_t$, and output gate $\mathbf{o}_t$. The key advantage of LSTM is its ability to remove useless information and store new knowledge through the memory cell. These behaviors are carefully controlled by the gates, which optionally let the information through. Let $\sigma(\cdot)$ and $\tanh(\cdot)$ be the sigmoid and the hyperbolic tangent functions, respectively. The computation process of LSTM can be expressed as below:
\vspace{-4pt}
\begin{equation}
   \begin{split}
      \mathbf{i}_t &= \sigma(\mathbf{W}_{i}\left[\mathbf{x}_t,\mathbf{h}_{t-1}\right] + \mathbf{b}_i) \\
      \mathbf{f}_t &= \sigma(\mathbf{W}_{f}\left[\mathbf{x}_t,\mathbf{h}_{t-1}\right] + \mathbf{b}_f) \\
      \mathbf{o}_t &= \sigma(\mathbf{W}_{o}\left[\mathbf{x}_t,\mathbf{h}_{t-1}\right] + \mathbf{b}_o) \\
      \mathbf{c}_t &= \mathbf{f}_t \odot \mathbf{c}_{t-1} + \mathbf{i}_t \odot \tanh(\mathbf{W}_{g}\left[\mathbf{x}_t,\mathbf{h}_{t-1}\right] +  \mathbf{b}_g) \\
      \mathbf{h}_t &= \mathbf{o}_t \odot \tanh(\mathbf{c}_t)
   \end{split}
   \label{eqn:LSTM}
\end{equation}
where $\mathbf{x}_t$ is the input at time-step $t$, and $\odot$ denotes the element-wise multiplication operation. The hidden state $\mathbf{h}_t$ can be fed to the softmax layer for prediction. We denote the computation process of equation (\ref{eqn:LSTM}) as $[\mathbf{h}_t, \mathbf{c}_t]=\mathrm{LSTM}(\mathbf{x}_t, \mathbf{h}_{t-1}, \mathbf{c}_{t-1})$ for notation simplification. 

\vspace{-8pt}

\section{Task representation}
In this section, we introduce the temporal And-Or graph (AOG), which captures rich  task-specific knowledge and enables to produce large amounts of valid training samples.

\vspace{-14pt}

\subsection{Temporal And-Or graph}
The AOG is defined as a 4-tuple set $\mathcal{G}=\{S, V_N, V_T, P\}$. $S$ is the root node denoting a task. The non-terminal node set $V_N$ contains both and-nodes and or-nodes. An and-node represents the decomposition of a task to its sub-tasks in chronological order. An or-node is a switch, deciding which alternative sub-task to select. Each or-node has a probability distribution $\mathbf{p}_t$ (the $t$-th element of $P$) over its child nodes, and the decision is made based on this distribution. $V_T$ is the set of terminal nodes. In our AOG definition, the non-terminal nodes refer to the sub-tasks and atomic actions, and the terminal nodes associate a batch of atomic actions. In this work, we manually define the structure of the AOG for each task.

According to this representation, the task ``pour water with the cup" can be represented as the AOG shown in Figure \ref{fig:aog}(a). The root node denotes the task, and it is first decomposed into two sub-tasks, i.e., ``grasp the cup" and ``pour water into the cup", under the temporal constraint. The ``grasp the cup" node is an and-node and can be further decomposed into ``move to the cup" and ``take the cup" in chronological order. The ``pour water into the cup" node is an or-node, and it has two alternative sub-branches, i.e., ``pour water with the water dispenser" and ``pour water with the pot". Finally, all the atomic actions are treated as the primitive actions and associated objects, which are represented by the terminal nodes. In this way, the temporal AOG contains all possible action sequences of the corresponding task in a syntactic way.

\vspace{-8pt}
\subsection{Sample generation with And-Or graph}

In addition to capturing the task semantics, the AOG representation enables to generate large amount of valid samples (i.e., action sequences extracted from the AOG), which are significant for the recurrent neural network learning. According to the definition of the AOG, a parsing graph, which corresponds to a specific action sequence (e.g., Figure \ref{fig:aog}(b)), will be generated by selecting the sub-branches for all the or-nodes searched in a DFS manner given a new scene image. Since the explicit temporal dependencies exist among these or-nodes, we can recurrently activate these selections by utilizing a LSTM network, i.e., AOG-LSTM.

\begin{figure}[!t]
   \centering
   \includegraphics[width=1.0\linewidth]{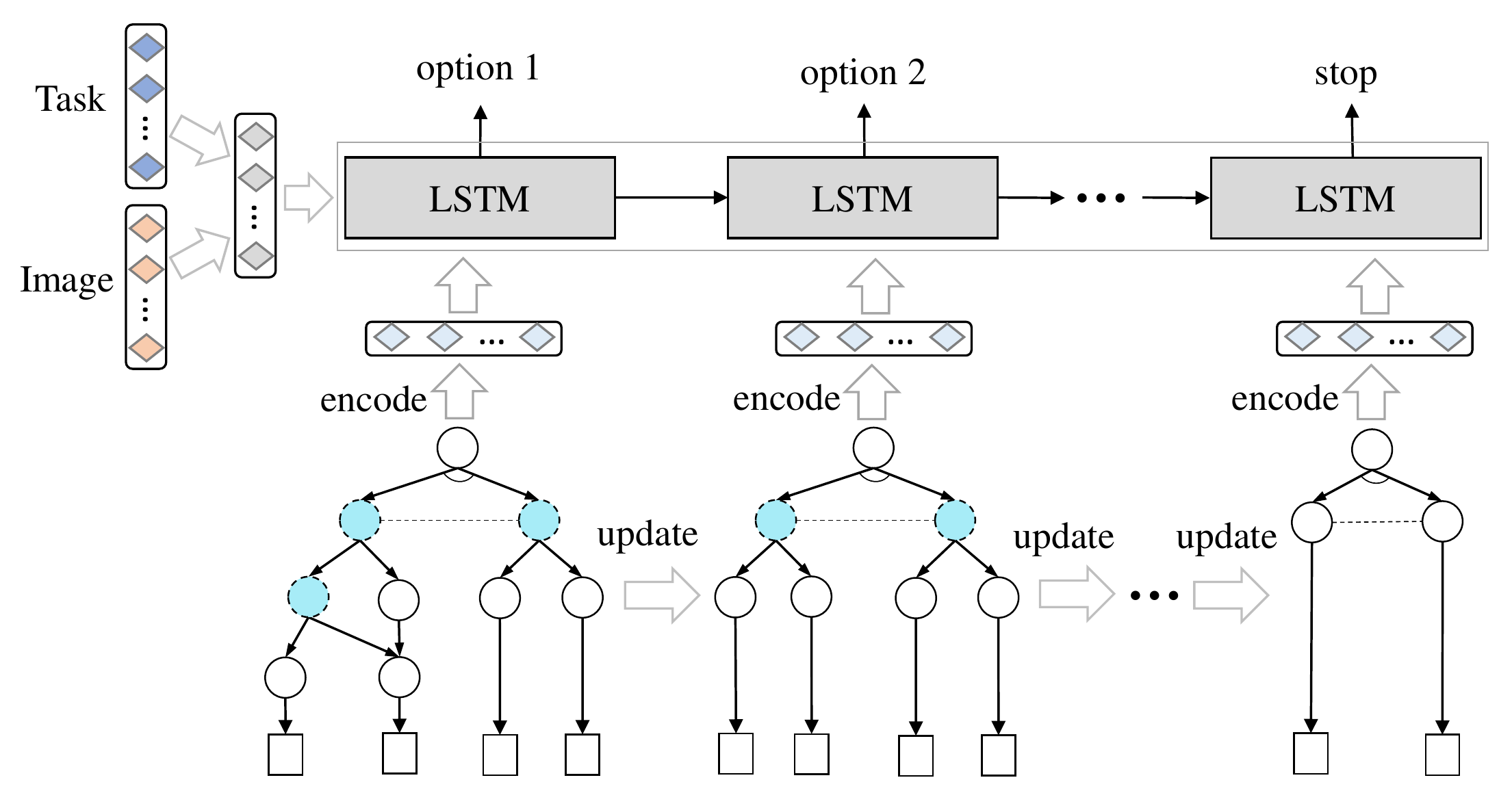}
      \vspace{-8pt}
   \caption{The architecture of AOG-LSTM. It is designed for selecting the sub-branches at all of the or-nodes in a temporal And-Or graph.}
   \label{fig:aog-lstm}
   \vspace{-8pt}
\end{figure}

As illustrated in Figure \ref{fig:aog-lstm}, our model first extracts the features of the given scene image and the task, and maps them to a feature vector, which serves as the initial hidden state of the AOG-LSTM. It then encodes the initial AOG as an adjacency matrix. The matrix is further re-arranged to be a vector, which is fed into the AOG-LSTM to predict the sub-branch selection of the first or-node. Meanwhile, the AOG is updated via pruning the unselected sub-branches. Note that the AOG is updated based on the annotated and predicted selection during training and test stages, respectively. Based on the updated AOG, the same process is conducted to predict the selection of the second or-node.
The process is repeated until all or-nodes have been visited, and a parsing graph is then constructed. We denote the image and task features as $\mathbf{f}^I$ and $\mathbf{f}^T$, the AOG features at time-step $t$ as $\mathbf{f}_t^{AOG}$. The prediction at time-step $t$ can be expressed as:

\vspace{-12pt}
\begin{equation}
   \begin{split}
   &\mathbf{c}_0 =  \mathbf{0}; \ \mathbf{h}_0=\mathbf{W}_{hf}[\mathbf{f}^I, \mathbf{f}^T] \\
   &[\mathbf{h}_t, \mathbf{c}_t] = \mathrm{LSTM}(\mathbf{f}^{AOG}_t, \mathbf{h}_{t-1}, \mathbf{c}_{t-1}) \\
   &\mathbf{p}_t = \mathrm{softmax}(\mathbf{W}_{hp}\mathbf{h}_t + \mathbf{b}_p)
   \end{split}
   \label{eqn:AOG-LSTM}
\end{equation}
where $\mathbf{p}_t$ is the probability distribution over all child branches of the $t$-th or-node, and the one with maximum value is selected. Making selection at the or-nodes is less ambiguous since the AOG representation effectively regularizes the semantic space. Thus, we can train the AOG-LSTM only using a small number of annotated samples. Specifically, we collect a small set of samples annotated with the selections of all or-nodes given a scene image for each task and define the cross-entropy objective function to train the AOG-LSTM.

Once the AOG-LSTM is trained,  we adopt it to predict the sub-branch selections for all the or-nodes in the And-Or graph given different scene images, and generate the corresponding action sequences. In this way, a relative large set of $(I,T,\mathcal{A})$ samples is obtained, where $I$, $T$, $\mathcal{A}$ represent the image, task and predicted sequence, respectively. More importantly, it can also generate samples of unseen tasks using identical process in which the AOG structures for the new tasks are also manually defined. These newly generated samples effectively alleviate the suffering of manually annotating large mounts of samples in practice.

\begin{figure}[!t]
   \centering
   \includegraphics[width=1.0\linewidth]{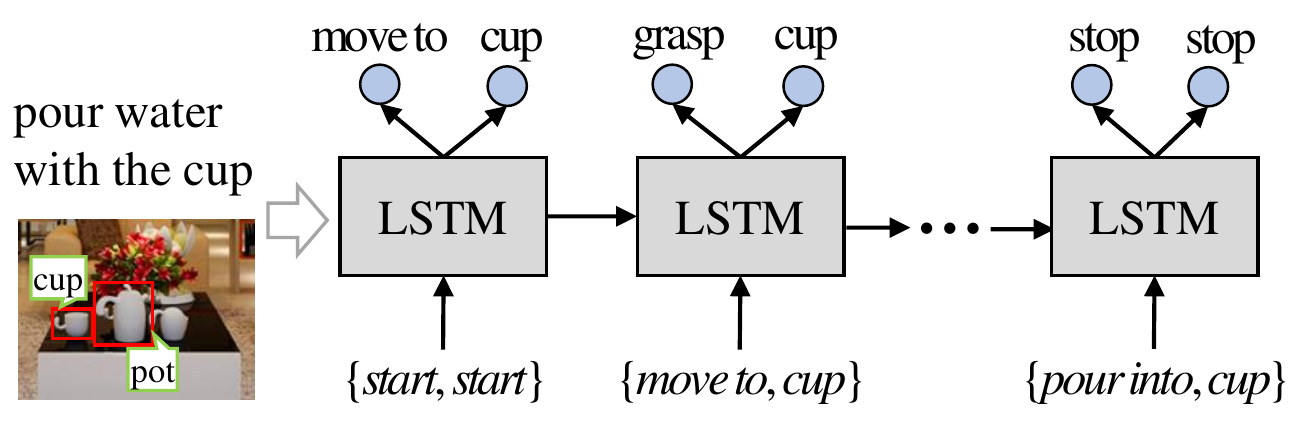} % requires the graphicx package
      \vspace{-4pt}
   \caption{The architecture of Action-LSTM for predicting the atomic action sequence given a specific task.}
   \label{fig:s-lstm}
   \vspace{-8pt}
\end{figure}

\section{Recurrent action prediction}

We formulate the problem of task-oriented action prediction in the form of probability estimation $p(A_1, A_2, ..., A_n|I,T)$, where $I$ and $T$ are the given scene image and the task, and $\{A_1, A_2, ..., A_n\}$ denotes the predicted sequence. Based on the chain rule, the probability can be recursively decomposed as:
\vspace{-10pt}
\begin{align}
   p(A_1, A_2, ..., A_n|I, T)=\prod_{t=1}^np(A_t|I, T, \mathcal{A}_{t-1}),
   \label{eqn:chainrule}
\end{align}
where $\mathcal{A}_{t-1}$ denotes $\{A_1, A_2, ..., A_{t-1}\}$ for convenient illustration. The atomic action is defined as $A_i=(a_i, o_i)$. In this work, we simplify the model by assuming the independence between the primitive actions and the associated objects. Thus the probability can be expressed as:
\vspace{-4pt}
\begin{align}
   p(A_t|I, T, \mathcal{A}_{t-1})=p(a_t|I, T, \mathcal{A}_{t-1})p(o_t|I, T, \mathcal{A}_{t-1}).
   \label{eqn:decomposition}
\end{align}

Here we develop the Action-LSTM network to model the probability distribution, i.e., equation (\ref{eqn:chainrule}).  Specifically, the Action-LSTM first applies the same process with AOG-LSTM to extract the features of the task and image, which is also used to initialize the hidden state of LSTM. At each time-step $t$, two softmax layers are utilized to predict the probability distributions $\mathbf{p}(a_t)$ over all primitive actions and $\mathbf{p}(o_t)$ over all associated objects. The conditions on the previous $t-1$ actions can be expressed by the hidden state $h_{t-1}$ and memory cell $c_{t-1}$. The action prediction at time-step $t$ can be computed by:
%\vspace{-1pt}
\begin{equation}
   \begin{split}
   &\mathbf{c}_0 =  \mathbf{0}; \ \mathbf{h}_0=\mathbf{W}_{hf}[\mathbf{f}^I, \mathbf{f}^T] \\
   &[\mathbf{h}_t, \mathbf{c}_t] = \mathrm{LSTM}(\mathbf{f}^A_t, \mathbf{h}_{t-1}, \mathbf{c}_{t-1}) \\
   &\mathbf{p}(a_t) = \mathrm{softmax}(\mathbf{W}_{ah}\mathbf{h}_t+\mathbf{b}_a) \\
   &\mathbf{p}(o_t) = \mathrm{softmax}(\mathbf{W}_{oh}\mathbf{h}_t+\mathbf{b}_o)
   \end{split}
   \label{eqn:AOG-LSTM}
\end{equation}
where $\mathbf{f}^A_t$ is the feature vector of the atomic action. Figure \ref{fig:s-lstm} gives an illustration of the Action-LSTM. At the training stage, we define the objective function as the sum of the negative log likelihood of correct sequence over the whole training set, including the manually-annotated and the automatically-generated samples, to optimize the network.

\begin{table*}[htbp]
\centering
\scriptsize
\begin{tabular}
{C{1.5cm}|p{0.412cm}p{0.412cm}p{0.412cm}p{0.412cm}p{0.412cm}p{0.58cm}p{0.442cm}|p{0.412cm}p{0.412cm}p{0.412cm}p{0.412cm}p{0.412cm}p{0.412cm}p{0.412cm}p{0.412cm}p{0.412cm}p{0.442cm}|p{0.412cm}p{0.412cm}}
\hline
\multirow{2}{*}{methods} &
\multicolumn{7}{c|}{primitive actions} &
\multicolumn{10}{c|}{associated objects} &
\multicolumn{2}{c}{average} \\
\cline{2-20}
& open  & grasp & put into & move to & pour into &put under& wash & bow & cup & pot  & water dis & tea box & ramen cup & ramen bag  & tap & basin & apple & action & object \\
\hline\hline
MLP & 94.50 & 90.70 & 93.20 & 96.60 & 82.20 & 64.30 & 84.80 & 95.20 & 90.90 & 96.30 & 99.20 & 97.70 & 95.90 & 100.00 & 96.00 & 64.30 & 95.50 & 86.60 & 93.10 \\
RNN & 95.70 & 94.10 & 95.10 & 95.70 & 88.50 & 64.00 & 87.90 & 96.40 & 94.50 & 95.10 & 98.40 & 98.50 & 96.30 & 100.00 & 89.90 & 63.60 & 93.90 & 88.60 & 92.70 \\
\hline
Ours w/o AOG & 94.00 & 94.30 & 93.70 & 96.70 & 92.50 & 63.60 & 100.00 & 99.30 & 95.00 & 96.30 & 98.40 & 98.10 & 97.40 & 100.00 & 97.00 & 81.80 & 100.00 & 90.70 & 96.30 \\
Ours w/ AOG & 95.80 & 95.10 & 94.40 & 96.80 & 94.60 & 63.60 & 100.00 & 99.80 & 95.70 & 96.40 & 98.40 & 98.10 & 98.10 & 100.00 & 96.00 & 81.80 & 100.00  & \textbf{91.50} & \textbf{96.40}\\
\hline
\end{tabular}
\caption{Accuracy results of recognizing atomic actions (i.e., primitive actions and associated objects) generated by our method with and without And-Or graph (Ours w/ and w/o AOG, respectively) and the two baseline methods (i.e., RNN and MLP).}
\label{table:primitive-results}
\end{table*}

\begin{table*}[htbp]
\centering
\scriptsize
\begin{tabular}
{C{1.5cm}|p{0.9cm}p{0.9cm}p{0.9cm}p{0.9cm}p{0.9cm}p{0.9cm}p{0.9cm}p{0.9cm}p{0.9cm}p{0.9cm}p{0.9cm}|p{0.9cm}}
\hline
methods & task1  & task2 & task3 & task4 & task5 & task6  & task7 & task8 & task9 & task10  & task11 & average\\
\hline\hline
MLP & 91.50 & 86.10 & 85.30 & 67.60 & 74.00 & 73.80 & 94.20 & 85.50 & 91.70 & 92.40 & 81.80  & 84.00 \\
RNN & 86.80 & 81.90 & 91.20 & 83.80 & 79.00 & 80.00 & 98.10 & 91.60 & 95.00 & 95.20 & 69.70  & 86.58 \\
\hline
Ours w/o AOG& 91.50 & 87.50 & 94.10 & 88.40 & 85.60 & 81.20 & 100.00 & 87.00 & 91.70 & 94.30 & 87.90 & 89.90 \\
Ours w/ AOG & 93.40 & 93.10 & 100.00 & 90.70 & 85.60 & 87.50 & 100.00 & 87.00 & 93.30 & 93.30 & 87.90 & \textbf{92.00} \\
\hline
\end{tabular}
\caption{Sequence accuracy results generated by our model with and without the And-Or graph (Ours w/ and w/o AOG) and the two baseline method (i.e., RNN and MLP).  We utilize task 1 to task 11 to denote tasks ``make ramen", ``make ramen in the ramen cup", ``make ramen in the ramen bag", ``pour water", ``pour water from the pot", ``pour water with the cup", ``pour water from water dispenser", ``make tea", ``make tea with the cup", ``make tea using water from the pot``, ``wash apple".}
\label{table:sequence-results}
\vspace{-12pt}
\end{table*}

\section{Experiments}
\subsection{Experimental setting}
\noindent\textbf{Dataset. }
To well define the problem of task-oriented action prediction, we create a large dataset that contains 13 daily tasks described by the AOGs and 861 RGB-D scene images. We define 4 general daily tasks, i.e., ``make tea", ``make ramen", ``pour water", ``wash apple", and 9 derived ones, i.e., ``pour water with cup", ``make tea using water from water-dispenser", etc. The derived tasks are similar to the general ones but with some required objects specified. The images are captured from 16 scenarios of 5 daily environments, i.e., lab, dormitory, kitchen, office and living room. All of the objects in these images are annotated with their class labels and location coordinates. As described above, the atomic action is defined as two-tuples, i.e., a primitive action and its associated object. In this dataset, we define 7 primitive actions, i.e., ``open", ``grasp", ``put into", ``move to", ``put under", ``pour into", ``wash", and 10 associated objects, i.e., ``bowl", ``cup", ``pot", ``water-dispenser", ``tea-box", ``ramen-cup", ``ramen-bag", ``tap", ``basin", ``apple".

The dataset includes three parts, i.e., training, testing sets and an augmented set generated from the AOGs. The training set contains 600 samples for 11 tasks with the annotation (i.e., the selections of all the or-nodes in the corresponding AOGs), and this training set is used to train the AOG-LSTM. The augmented set contains 2,000 samples of ${(I, T, \mathcal{A}^p)}$ , in which $\mathcal{A}^p$ is the predicted sequence. For training the Action-LSTM, we combine the augmented set and training set as a whole. The testing set contains 1,270 samples of ${(I, T, \mathcal{A})}$ for performance evaluation.

\noindent\textbf{Implementation details. }
$\mathbf{f}^I$ is the feature vector containing the class labels and locations of the objects in image $I$, and $\mathbf{f}_t^A$ is a one-hot vector denoting a specific task. $\mathbf{f}^T$ is concatenated by two one-hot vectors, denoting the primitive action and object, respectively.  We implement our models using the Caffe framework \cite{jia2014caffe}, and train the AOG-LSTM and Action-LSTM using the SGD with momentum of 0.9, weight decay of 0.0005, batch size of 40, and initial learning rates of 0.01.

\begin{figure}[htbp]
\centering
\subfigure[]{
\includegraphics[width=0.47\linewidth]{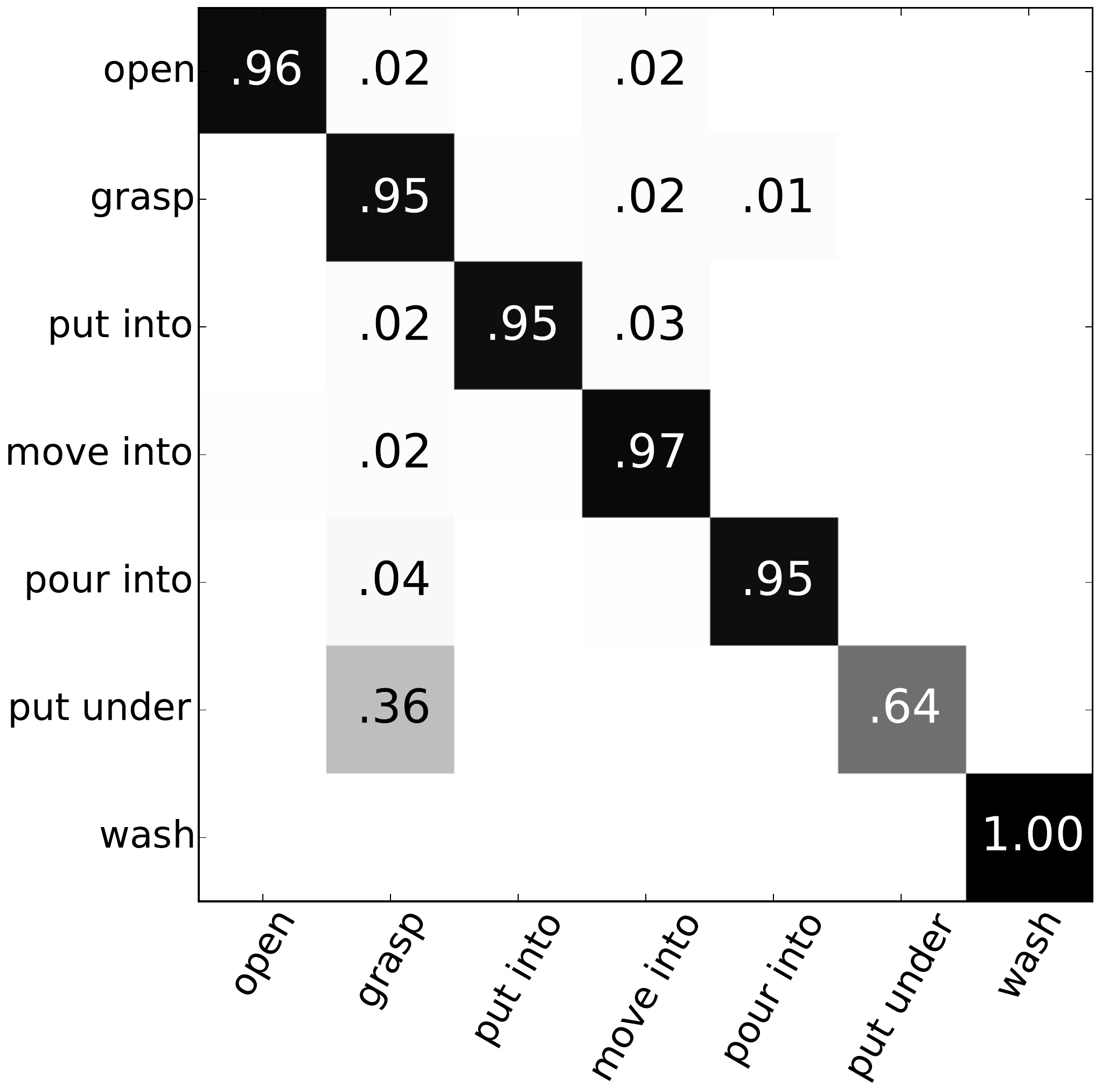}}
\subfigure[]{
\includegraphics[width=0.47\linewidth]{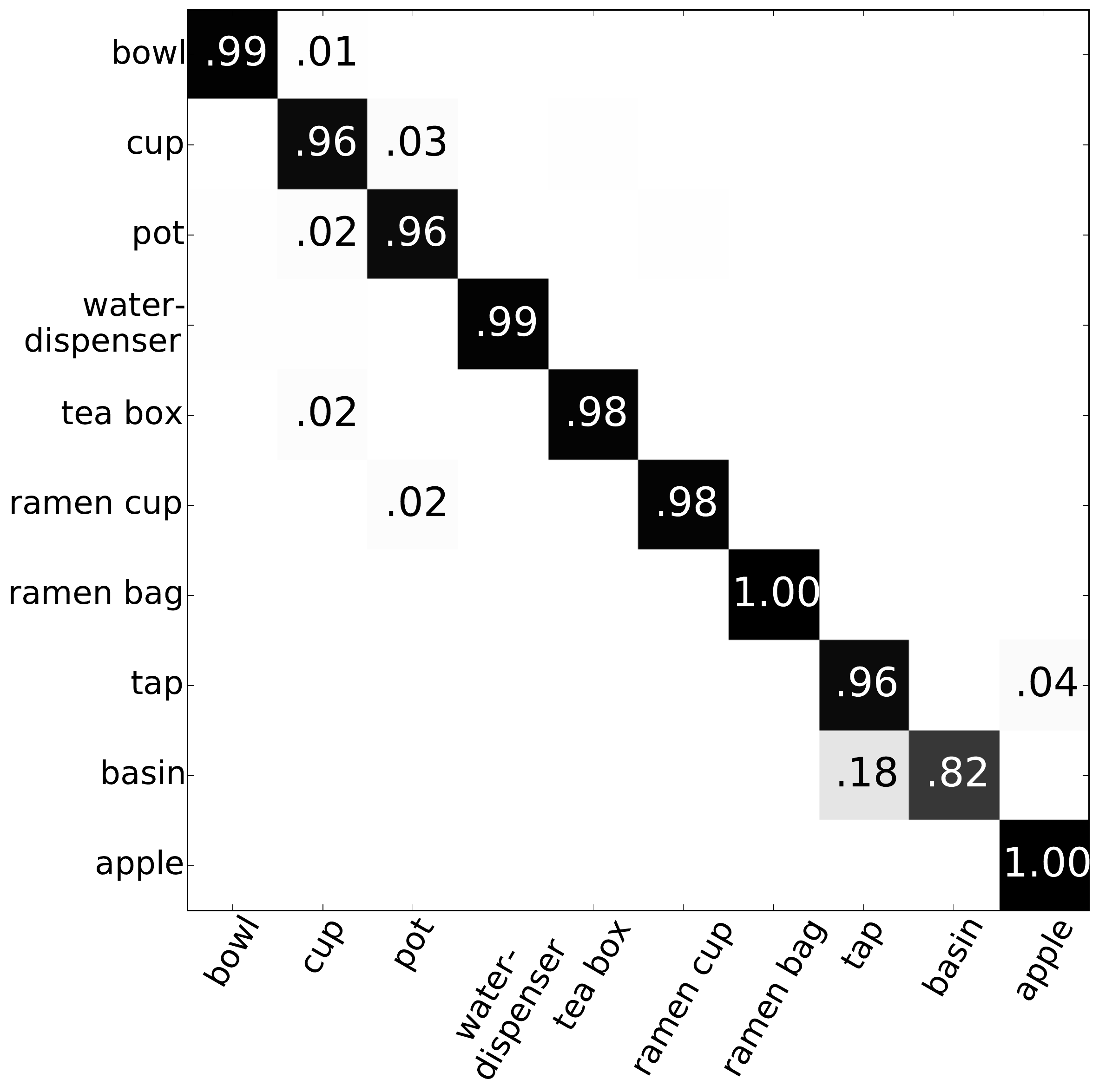}}
\vspace{-12pt}
\caption{The confusion matrixes for (a) 7 primitive actions and (b) 10 associated objects of our model.}
\label{fig:confuse-matrix}
\end{figure}

\vspace{-14pt}

\subsection{Results and analysis}
\subsubsection{Comparisons with baseline models}
To verify the effectiveness of our model, we implement two neural networks as the baselines for this task, i.e., multi-layer perception (MLP) and recurrent neural network (RNN).  The MLP predicts the $t$-th atomic action by taking the task features and image features, and the previous $t-1$ predicted atomic actions as input. It repeats the process until a stop signal is obtained. The learning and inference processes of RNN is exactly the same with training our Action-LSTM. For fair comparison, all the networks have one hidden layer of 512 neurons.

We first evaluate the performance of our model for recognizing atomic actions. Figure \ref{fig:confuse-matrix} presents the confusion matrixes for the predicted primitive actions and associated objects, where our model achieves very high accuracies on most of the classes. Table \ref{table:primitive-results} further depicts the detailed comparison of our model against the baseline methods. Our model can predict the primitive actions and associated objects with the accuracies of 96.40\% and 91.50\% on average, outperforming the baseline methods by 2.9\% and 3.3\%.

Then we evaluate the accuracy of generating the action sequences, i.e., whether the task is completed successfully. We define the sequence accuracy as the fraction of complete correct sequences with respect to all predicted sequences. The results on the sequence accuracy are reported in Table \ref{table:sequence-results}. Our model can correctly predict complete action sequences with a probability of 92\% on average. It evidently outperforms the baseline methods on most of the tasks (8/11) by improving 5.42\% accuracy on average.

\vspace{-10pt}

\subsubsection{Benefit of using And-Or graph}
In this experiment, we empirically evaluate the contribution of introducing AOG for the neural network learning. Here we train the Action-LSTM with and without using the augmented sample set, and report the results in the last two rows of Table \ref{table:primitive-results} and Table \ref{table:sequence-results}, i.e., Ours w/ and w/o AOG.
It can be observed that the results using AOGs have a notable improvement on both atomic action recognition and sequence prediction. The performance improvements clearly demonstrate the effectiveness of adopting the augmented set. In particular, generating samples from AOG representations enable us to better capture the complex task variations and it is an effective way to compensate the neural network learning. Besides, it is noteworthy that the Action-LSTM performs better than traditional RNN model, since LSTM has better ability for memorizing the long term dependencies among actions. 

\vspace{-10pt}

\subsubsection{Generalization to related tasks}

Here we ``related tasks'' as the ones that have similar atomic actions or temporal context with  the existing tasks in the training set. For example, ``pour water with the bowl" is a related task to ``pour water". Then, it would be interesting to see how our trained model can be generalized to the related tasks. In particular, for each related task, we have its AOG representation but no annotated training samples. In this experiment, the models of MLP, RNN and Ours without AOG are all trained on the training set, which only contains the samples of task 1 to task 11, as described above. For our model with AOG, we first train the AOG-LSTM with the same set of the annotated samples as the other competing models. Then we utilize the trained AOG-LSTM to produce samples for all tasks, including task 12 and task 13, and then use these samples to train the Action-LSTM. The results of two tasks are presented in Table \ref{table:generlization}. We find that the performances of the three methods without using AOG are  extremely unsatisfying on the both tasks. By comparison, our approach with the AOG representations boosts the average sequence accuracy to 73.05\%, outperforming others by nearly 50\%. These results well demonstrate the excellent generalization ability of our model.

\begin{table}[!t]
\centering
\scriptsize
\begin{tabular}{c|ccc}
\hline
methods & task 12  & task 13 & average\\
\hline\hline
MLP & 34.10 & 0.00 &  17.05\\
RNN & 36.60 & 0.00 & 18.30 \\
\hline
Ours w/o AOG& 41.50 & 0.00  & 24.40\\
Ours w/ AOG & 82.90 & 63.20 & \textbf{73.05} \\
\hline
\end{tabular}
\vspace{-4pt}
\caption{Sequence accuracy results generated by our model with and without And-Or graph (Ours w/ and w/o AOG) and two baseline method (i.e., RNN, MLP). Task 12 and task 13 denote tasks ``make tea using water from the water dispenser", ``pour water with the bowl", respectively.}
\label{table:generlization}
\vspace{-10pt}
\end{table}

\vspace{-6pt}
\section{Conclusion}

In this paper, we address a challenging problem, i.e., predicting a sequence of actions to accomplish a specific task under a certain scene, by developing a recurrent LSTM neural network. To alleviate the issue of requiring large amounts of annotated data, we present a two-stage model training approach by employing a temporal And-Or graph representation. From this representation, we can produce a large number of valid samples (i.e., task-oriented action sequences) that facilitate the LSTM network learning. Extensive experiments on a newly created dataset demonstrate the effectiveness and flexibility of our approach. In future work, we will explore to automatically learn the And-Or graph structure and generalize our approach to other intelligent applications.

% References should be produced using the bibtex program from suitable
% BiBTeX files (here: strings, refs, manuals). The IEEEbib.bst bibliography
% style file from IEEE produces unsorted bibliography list.
% -------------------------------------------------------------------------
\bibliographystyle{IEEEbib}
\small
\bibliography{reference}

\end{document}